\documentclass[11pt]{article}

\usepackage[a4paper,margin=1in]{geometry}
\usepackage[T1]{fontenc}
\usepackage[utf8]{inputenc}
\usepackage{lmodern}
\usepackage{microtype}
\usepackage{amsmath,amssymb}
\usepackage{booktabs}
\usepackage{tabularx}
\usepackage{multirow}
\usepackage{xcolor}
\usepackage{graphicx}
\usepackage{enumitem}
\usepackage{fancyhdr}
\usepackage{tcolorbox}
\usepackage{url}
\usepackage[hidelinks]{hyperref}
\usepackage[capitalize,noabbrev]{cleveref}

\newcommand{\model}[1]{\texttt{#1}}
\newcommand{\dataset}[1]{\textsc{#1}}

\newcommand{\stagebox}[2]{%
  \fcolorbox{blue!35}{blue!4}{%
    \parbox[c][3.05cm][c]{0.185\linewidth}{\centering\textbf{#1}\\[0.3em]\footnotesize #2}}}

\title{Data-Centric Post-Training for Financial Reasoning:\\
Mining, Distillation, and Verifiable Learning\\
\large Technical Report}
\author{Zhirayr Hayrapetyan, Andrei Kalmykov, Denis Kokosinskii,\\
Dmitry Stanishevskii, Dmitry Zmitrovich}
\date{}

\begin{document}
\maketitle

\begin{tcolorbox}[
  colback=cyan!5,
  colframe=cyan!5,
  boxrule=0pt,
  arc=2mm,
  left=3mm,right=3mm,top=2.5mm,bottom=2.5mm]
\small
\textbf{Abstract.}
Financial text, textbooks, and question--answer pairs are abundant, but only a small fraction is directly usable for reasoning-focused post-training. Existing QA pairs often lack explicit reasoning, sufficient context, or reliably verifiable answers, while textbooks must first be transformed into synthetic training examples. We present a data-centric pipeline that constructs complementary corpora by mining open-source reasoning traces, distilling financial instruction data, and generating knowledge-graph-guided question--answer pairs from financial educational material. After semantic deduplication, three lightweight sequence classifiers select finance-relevant examples, reject under-specified questions, and identify tasks suitable for reinforcement learning with compact rule-based verifiers. For model adaptation, we study supervised fine-tuning and reinforcement learning, while self-distilled fine-tuning and post-training model merging are used to prevent the loss of financial capabilities already present in the starting model. We evaluate the adapted language models using FINESSE-Bench, reporting aggregate performance and changes relative to their starting checkpoints. Across the selected comparisons, ordinary SFT reduces FINESSE-Bench accuracy by 3.2--4.0 percentage points, whereas self-distilled SFT improves over the corresponding starting models by 1.0--2.8 points. Equal-weight merging recovers 3.0 points over its SFT parent and finishes 0.9 points above the original model; GRPO on hard tasks adds 0.4 points after self-distilled SFT or 3.0 points when applied directly to verifiable tasks. These results show that retention-aware adaptation can improve financial reasoning without the regressions observed after ordinary SFT.
\end{tcolorbox}

\section{Introduction}
\label{sec:introduction}

Large language models (LLMs) are increasingly capable of natural language understanding, general instruction following and multi-step reasoning, yet their deployment in vertical domains, particularly finance, remains challenging. A useful response may require specialized terminology, numerical accuracy, evidence integration across text and tables, and a conclusion that can be audited. These requirements expose two weaknesses of naive domain adaptation. First, existing public financial instruction datasets provide limited coverage of specialized concepts and reasoning patterns; available examples may also contain repeated templates, missing context, and answers of uneven quality. This motivates both mining additional reasoning traces and synthesizing questions from financial educational material. Second, straightforward fine-tuning on a narrow dataset can improve performance on some financial tasks while degrading financial capabilities already present in the starting model---a form of catastrophic forgetting.

Recent work on financial reasoning shows that domain adaptation depends not only on the training algorithm but also on how examples are constructed, verified, and selected for each post-training stage~\cite{liu2025finr1,qian2025fino1,cao2026odafin}. Our work develops a complementary production-oriented pipeline around the same central observation: different post-training stages require different notions of data quality, and these notions should be modeled explicitly rather than collapsed into a single opaque quality score.

We organize data curation around three binary decisions. The \emph{finance relevance} decision supports high-recall mining from large general reasoning collections. The \emph{missing-context} decision removes prompts that cannot be answered from the information presented, a pathology that is especially common in synthetic data. The \emph{verifiability} decision identifies questions whose answers admit compact numerical, multiple-choice, or pattern-based checking and are therefore suitable for RL with rule-based rewards. Each detector is implemented as a sequence-classification head over a compact Qwen3-Embedding backbone~\cite{zhang2025qwen3embedding}, allowing hundreds of thousands of candidates to be screened more economically than with a generative judge.

The curated corpus has three complementary components. First, we mine long financial reasoning traces from \dataset{Glaive Reasoning v1 20M}~\cite{glaive_reasoning}. Second, we distill and filter \dataset{Finance-Instruct-500K}~\cite{finance_instruct}, retaining a set of finance-relevant, self-contained examples with explicit reasoning. Third, we use GraphGen~\cite{chen2025graphgen} to extract structured knowledge from financial educational material and generate diverse question--answer pairs from selected parts of the resulting knowledge graph, ranging from questions about individual facts to questions that require combining related concepts. Across all sources, semantic deduplication, language filtering, repeated-prefix removal, and classifier thresholds reduce both redundancy and synthetic-data artifacts.

Our eventual experimental analysis considers four adaptation mechanisms: conventional SFT, SFT with self-distillation, model merging, and RL. The first two inject domain behavior directly; merging interpolates adapted and starting-model parameters to recover financial capabilities lost during fine-tuning; and RL goes beyond imitation by sampling alternative responses and optimizing them against outcome-level reward signals. This report focuses on the data and training decisions that make those comparisons interpretable.

The report makes the following contributions:
\begin{itemize}[leftmargin=1.5em]
    \item We describe a scalable, classifier-assisted mining and filtering pipeline in which finance relevance, self-containment, and verifiability serve separate operational purposes.
    \item We construct three complementary financial post-training sets: 44,652 mined long-reasoning examples, 4,475 distilled Finance-Instruct examples, and 329,828 knowledge-graph-guided synthetic examples.
    \item We document a GraphGen adaptation for financial educational material and a multi-stage filtering pipeline that combines semantic, lexical, language, difficulty, and learned signals.
    \item We show that self-distillation and post-training merging both mitigate the loss of financial capability caused by ordinary SFT. In the selected Qwen3.5-4B results, self-distillation is more effective, finishing 2.8 FINESSE-Bench points above the starting model compared with 0.9 points for merging.
\end{itemize}

\subsection{Related Work}
\label{sec:related}

\paragraph{Financial reasoning models.}
Financial LLM research has increasingly moved from domain vocabulary and sentiment classification toward multi-step numerical and analytical reasoning. Fin-R1 constructs a financial chain-of-thought corpus and applies SFT followed by RL~\cite{liu2025finr1}. Fino1 evaluates the transfer of general reasoning enhancements to financial text, table, and equation tasks and reports that chain-of-thought adaptation and RL provide non-uniform gains across task types~\cite{qian2025fino1}. ODA-Fin takes a data-centric view, combining semantic deduplication, reasoning synthesis, length-adaptive verification, and difficulty-aware selection for RL~\cite{cao2026odafin}. These works motivate our emphasis on explicit data profiles and stage-specific filtering.

\paragraph{Financial model evaluation.}
FINESSE-Bench provides a hierarchical evaluation of financial competence across CFA exam questions, technical analysis, and derivatives trading~\cite{stanishevskii2026finesse}. We use FINESSE-Bench to evaluate the target language models.

\paragraph{Reasoning distillation and verifiable RL.}
Reasoning traces from stronger models provide dense supervision for smaller or less specialized models. DeepSeek-R1 demonstrates the effectiveness of distilling long reasoning behavior and using rule-based rewards on domains with checkable answers~\cite{deepseekai2025r1}. GRPO removes the need for a learned value model by normalizing rewards within groups of sampled responses~\cite{shao2024deepseekmath}. In finance, however, many questions are open-ended or depend on missing documents. Applying rule-based RL indiscriminately can therefore reward formatting rather than correctness. Our verifiability and missing-context classifiers explicitly identify the subset on which compact rewards are defensible.

\paragraph{Knowledge-guided synthetic data.}
Synthetic question generation can expand domain coverage but often produces repetitive, under-specified, or factually weak tasks. GraphGen first segments source documents and uses a synthesis model to extract entities, relations, and their descriptions into a fine-grained knowledge graph~\cite{chen2025graphgen}. It then organizes selected knowledge into constrained subgraphs and uses the same model to convert those subgraphs into question--answer pairs. This process supports questions about individual facts as well as questions that combine or connect several related concepts. We apply this workflow to financial educational material. Because GraphGen does not produce explicit reasoning traces, we subsequently run a separate reasoning model over the generated questions and retain enriched examples whose derived answers pass answer verification.

\paragraph{Deduplication and efficient filtering.}
Exact-string matching removes literal duplicates but misses paraphrased templates and near-identical synthetic tasks. We therefore combine embedding-based semantic deduplication with deterministic normalization and common-prefix filtering. Qwen3-Embedding models provide multilingual semantic representations at several parameter scales~\cite{zhang2025qwen3embedding}; we use the 8B variant for semantic deduplication and the 0.6B backbone for the three binary filters.

\section{Data Construction}
\label{sec:data}

\subsection{Pipeline Overview}

We construct three complementary data branches from Glaive Reasoning, Finance-Instruct, and financial books processed with GraphGen. In each branch, semantic deduplication is performed at the prompt level. We then remove exact duplicates after text normalization and suppress groups of prompts with unusually long shared prefixes. English-language checks and the three auxiliary classifiers serve explicit purposes: the finance classifier selects domain-relevant examples, the missing-context classifier rejects questions that cannot be answered from the provided information, and the verifiability classifier identifies examples suitable for compact rule-based rewards.

\begin{figure*}[t]
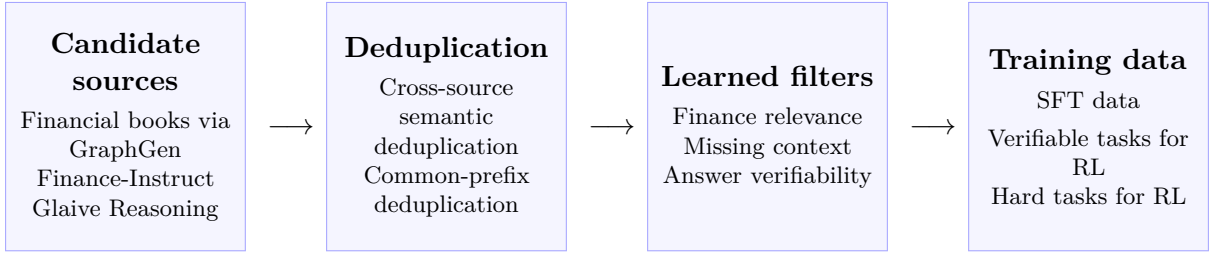

\centering
\stagebox{Candidate sources}{Financial books via GraphGen\\Finance-Instruct\\Glaive Reasoning}
\hfill$\longrightarrow$\hfill
\stagebox{Deduplication}{Cross-source semantic\\deduplication\\Common-prefix\\deduplication}
\hfill$\longrightarrow$\hfill
\stagebox{Learned filters}{Finance relevance\\Missing context\\Answer verifiability}
\hfill$\longrightarrow$\hfill
\stagebox{Training data}{SFT data\\[0.35em]Verifiable tasks for RL\\Hard tasks for RL}
\caption{Overview of the data-centric post-training pipeline.}
\label{fig:data-pipeline}
\end{figure*}

\begin{table}[t]
\centering
\small
\caption{Final curated datasets}
\label{tab:data-summary}
\begin{tabularx}{\linewidth}{@{}l r X@{}}
\toprule
Dataset & Examples & Primary construction route\\
\midrule
\dataset{CFA-GraphGen} & 329,828 & Knowledge-graph-guided synthesis from financial educational material \\
\dataset{Glaive-Finance} & 44,652 & Finance mining from \dataset{Reasoning-v1-20M} followed by strict filtering \\
\dataset{Finance-Instruct-Distilled} & 4,475 & Domain filtering and answer/reasoning distillation from \dataset{Finance-Instruct-500K} \\
\bottomrule
\end{tabularx}
\end{table}

\paragraph{Released data.}
We release the scored candidate pools as \href{https://huggingface.co/datasets/whoisjiji/fin-glaive}{\dataset{Fin-Glaive}} (645,232 examples) and \href{https://huggingface.co/datasets/whoisjiji/fin-cfa-graphgen}{\dataset{Fin-CFA-GraphGen}} (785,149 examples). These releases expose the financial-relevance, self-containment, and answer-verifiability scores before the stricter thresholds used to construct the training subsets in \cref{tab:data-summary}, allowing users to select thresholds for their own applications. The Finance-Instruct-derived component is not redistributed.

\subsection{Shared Semantic Deduplication}

All three sources undergo prompt-level semantic deduplication using the 8B Qwen3-Embedding model~\cite{zhang2025qwen3embedding} in a NeMo Curator workflow~\cite{nemo_curator}. Candidate embeddings are partitioned into 100 clusters; within-cluster pairs use a cosine-distance threshold $\epsilon=0.1$. The workflow retains representatives closest to the cluster centroid. We also remove families of prompts with unusually long shared prefixes, a common artifact of synthetic or template-driven generation in which examples differ only in a few entities or numbers.

\subsection{Mining Financial Reasoning from Glaive}
\label{sec:glaive}

We use the finance detector described in \cref{sec:classifiers} to mine candidate prompts from the 20-million-example \dataset{Glaive Reasoning v1} collection~\cite{glaive_reasoning}. The initial mining pass yields approximately 700,000 finance candidates. After semantic deduplication, response parsing, and exact prompt deduplication, 645,232 examples remain.

We deliberately apply stringent thresholds because manual inspection found that lower finance-confidence cohorts contained substantial adjacent-domain or generic reasoning content. Requiring $p(\mathrm{finance})>0.99$ reduces the pool to 220,707. Requiring self-containment probability above 0.9 leaves 111,655. Normalized exact deduplication leaves 111,643, and repeated-prefix filtering leaves 93,570. We retain only examples for which the prompt, reasoning, and final answer are classified as English by fastText language identification~\cite{joulin2016fasttext}, and separately reject strings containing Chinese characters; 93,427 survive these checks. Finally, we require at least 1,000 tokens in the reasoning trace under the Qwen3.5 tokenizer. The resulting \dataset{Glaive-Finance} corpus contains 44,652 long-reasoning examples.

\subsection{Distilling Finance-Instruct}
\label{sec:finance-instruct}

The Finance-Instruct branch first retains examples for which both the user prompt and answer are classified as English and removes examples containing Chinese characters. We then apply prompt-level semantic deduplication; the resulting materialized pool contains 287,572 examples from \dataset{Finance-Instruct-500K}~\cite{finance_instruct}. We manually map the dataset's system prompts to task families and remove non-generative families, including classification, named-entity annotation, and other structured labeling tasks, leaving 214,323 examples.

We next apply the financial-domain detector at $p(\mathrm{finance})>0.5$, leaving 79,175 examples. Common-prefix filtering and joint English-language checks reduce the pool to 68,312.
Requiring self-containment probability above 0.9 leaves 20,764 examples. We then apply the same financial-domain detector more stringently, requiring $p(\mathrm{finance})>0.99$ and leaving 11,809 examples. We retain only parsable conversation structures, then restrict the current training artifact to single-turn examples. This yields 4,475 examples. The retained traces were generated by DeepSeek-R1-0528~\cite{deepseekai2025r1} with temperature 0.6, top-$p$ 0.95, high reasoning effort, and a maximum completion length of 32,768 tokens.

\subsection{Knowledge-Graph-Guided Synthesis from Financial Material}
\label{sec:graphgen}

Our largest component adapts GraphGen~\cite{chen2025graphgen} to CFA curriculum and exam-preparation books covering core areas such as economics, financial analysis, equity, fixed income, derivatives, alternative investments, portfolio management, and professional ethics. Documents are divided into 1,024-token chunks with an overlap of 100 tokens. We use Qwen2.5-72B-Instruct~\cite{yang2024qwen25} both to extract entities and relations into a fine-grained knowledge graph and to synthesize questions and answers from that graph. Generation uses temperature 0, top-$p$ 0.95, and a maximum of 8,192 output tokens.

The local generation pool contains 785,149 unique prompts after semantic deduplication. Finance-confidence filtering at 0.99 leaves 706,699 examples, and a self-containment threshold of 0.9 leaves 440,183. Normalized exact deduplication and common-prefix removal reduce this to 335,158. Joint English detection over prompt and answer, followed by Chinese-character removal, leaves 331,458 examples.

To obtain explicit reasoning traces for model training, we seed-shuffle this curated pool and select 40,000 questions. Each question is submitted by itself to Kimi K2.6~\cite{moonshot2026kimik26}, without revealing the original GraphGen answer, using temperature 1.0 and top-$p$ 1.0. The API returns the model's reasoning separately from its visible response; we store these as the assistant's reasoning trace and final answer, respectively. A separate GPT-5.5 judge compares the final answer with the original GraphGen reference and retains only factually consistent responses without material omissions or contradictions. The resulting verified artifacts contain 32,000 examples.

\subsection{Why the Three Sources Are Complementary}

The three branches trade scale for control. GraphGen supplies broad concept coverage and structured compositional questions. Glaive contributes naturally diverse, long reasoning trajectories mined from a much larger general pool. Finance-Instruct contributes a smaller set whose task type, provenance, reasoning, and final response were jointly curated.

\section{Auxiliary Data Classifiers}
\label{sec:classifiers}

Generative LLM judges are flexible but expensive to apply to millions of candidates. We therefore distill three recurring curation decisions into compact binary sequence classifiers: financial relevance, missing context, and answer verifiability. Each model adds a two-class classification head to \model{Qwen/Qwen3-Embedding-0.6B}~\cite{zhang2025qwen3embedding} and consumes the user question truncated to 4,096 tokens where this setting is preserved. The classifiers produce continuous positive-class probabilities so that thresholds can be selected separately for mining, conservative filtering, and RL data selection.

\paragraph{Released checkpoints.}
The checkpoints and model cards are available on Hugging Face:
\begin{itemize}[nosep,leftmargin=*]
    \item financial relevance: \url{https://huggingface.co/whoisjiji/fin-relevance};
    \item missing context: \url{https://huggingface.co/whoisjiji/lack-of-context-detector};
    \item answer verifiability: \url{https://huggingface.co/whoisjiji/verifiability-classifier}.
\end{itemize}
Each card provides label definitions, validation metrics, inference examples, and threshold-selection guidance.

\begin{table}[h!]
\centering
\small
\caption{Auxiliary classifiers and their saved validation events. F1, precision, and recall are for the positive class.}
\label{tab:classifiers}
\begin{tabularx}{\textwidth}{@{}>{\raggedright\arraybackslash}p{0.18\textwidth} >{\raggedright\arraybackslash}X r r r r@{}}
\toprule
Classifier & Positive class and downstream purpose & F1 & Accuracy & Precision & Recall \\
\midrule
Financial relevance & Finance-relevant question; mine domain examples from broad instruction/reasoning corpora & 0.97 & 0.97 & 0.98 & 0.96 \\
Missing context & Missing context; reject questions that cannot be answered from the provided information & 0.94 & 0.94 & 0.98 & 0.90 \\
Answer verifiability & Compactly verifiable answer; select tasks for deterministic or regular-expression rewards & 0.95 & 0.95 & 0.98 & 0.92 \\
\bottomrule
\end{tabularx}
\end{table}

\subsection{Financial Relevance}

The finance classifier is a mining model rather than a benchmark topic classifier. Its positive class includes banking, investments, insurance, taxes, cryptocurrency, financial planning, corporate finance, economic theory, and personal finance; the negative class is non-financial content.

To construct the training corpus, we combine 282,400 prompts sampled from Finance-Instruct with 165,623 prompts sampled from the synthetic GraphGen pool, giving 448,023 unique prompts. Qwen2.5-72B-Instruct assigns each prompt a multiclass financial-domain label, which we collapse to a binary finance target. The resulting dataset contains 443,336 labeled examples. We sample 2,500 examples per class for a balanced 5,000-example holdout, leaving 438,336 training examples with 57.9\% positive labels.

Training uses one epoch, BF16, per-device batch size 4, 32 gradient-accumulation steps, learning rate $10^{-5}$, fused AdamW, 0.01 weight decay, cosine decay with 10\% warmup, and maximum gradient norm 10. The saved best event occurs at step 1,600 with positive-class F1 0.97.

\subsection{Missing Context}

Synthetic questions frequently refer to absent tables, documents, entities, or earlier turns. Training directly on such examples teaches a model either to hallucinate the missing premise or to imitate unsupported answers. The missing-context classifier therefore predicts whether the presented question can be answered from the information it contains. The positive target denotes a question with missing context, and downstream pipelines use one minus the missing-context probability as a self-containment score.

To construct the training corpus, we combine 116,221 Finance-Instruct prompts with approximately 145,000 synthetic GraphGen prompts. GPT-4.1-mini labels whether each question has sufficient context to be answered as written. After response parsing and prompt deduplication, the dataset contains 259,432 labeled examples. We reserve 2,500 examples from each class and retain 254,430 training examples. Hyperparameters match the financial-relevance classifier; the saved best event is step 300 with positive-class F1 0.94.

\subsection{Answer Verifiability}

The verifiability classifier separates questions with compact, objectively checkable outcomes from open-ended analysis. Positive examples include numerical calculations, multiple-choice questions, and tasks whose final answer can be normalized and checked by a small parser or regular expression. This classifier does not judge whether a supplied answer is correct; it judges whether correctness can be evaluated reliably with a compact verifier. The distinction is central to RL: a precise reward on a narrower task set is preferable to a noisy reward that mistakes stylistic overlap for correctness.

The deployed checkpoint uses the same Qwen3-Embedding-0.6B sequence-classification architecture. Hyperparameters match the financial-relevance classifier; the saved best event occurs at step 200 with positive-class F1 0.95.

\section{Training Strategies}
\label{sec:training}

The curated data supports four complementary adaptation strategies. Their shared objective is not only to increase financial capability, but also to control the loss of financial behavior already present in the starting model. We therefore treat domain gain and financial-capability retention as joint objectives throughout the experimental design.

\subsection{Supervised Fine-Tuning}

Conventional SFT provides the direct domain-adaptation baseline. Examples are represented as conversations with explicit reasoning and final-answer fields where available. GraphGen, mined Glaive reasoning, and distilled Finance-Instruct examples are normalized to this shared schema and combined into the SFT training mixture.

We use the same SFT optimization recipe across model families and for both ordinary and self-distilled SFT. Runs train for one epoch with global batch size 32, local batch size 1, BF16, FSDP2, and activation checkpointing. We update the language-model parameters only; for multimodal backbones, the vision and audio towers remain frozen. The optimizer is AdamW with $(\beta_1,\beta_2)=(0.9,0.95)$, $\epsilon=10^{-8}$, learning rate $10^{-5}$, weight decay 0.01, and cosine decay to $10^{-7}$.

\subsection{Self-Distilled SFT}

Kaplan et al.~\cite{kaplan2026hallucinations} characterize SFT-induced hallucinations as factual forgetting caused by interference among overlapping semantic representations. They show that self-distillation mitigates this interference by constraining drift from the model's earlier output distribution while still allowing new facts to be learned. Motivated by this result, we use a frozen copy of the starting checkpoint as the teacher during domain SFT.

For self-distilled SFT, the student and frozen teacher are initialized from the same starting checkpoint. The loss assigns weight 0.8 to knowledge distillation and 0.2 to supervised cross-entropy; distillation uses temperature 1.0 and computes the loss in FP32 chunks of 512 tokens. Ordinary SFT uses the otherwise identical training recipe without the teacher or distillation term.

\subsection{Model Merging}

As a post-training alternative, we linearly interpolate an SFT checkpoint with its starting checkpoint. This follows the weight-space interpolation strategy of WiSE-FT~\cite{wortsman2022wiseft}, which recovers capabilities lost during fine-tuning without adding inference-time cost. The selected experiment uses equal weights for the two checkpoints.

\subsection{Reinforcement Learning}

We use Group Relative Policy Optimization (GRPO)~\cite{shao2024deepseekmath} in two settings. In the first, following ODA-Fin's model-relative hard-example selection~\cite{cao2026odafin}, we revisit the questions in the same training dataset used for SFT. After completing SFT, we use the resulting checkpoint to generate four responses to each question and retain questions for which at least three responses are judged incorrect. We then continue training this checkpoint with GRPO on this hard subset of the SFT training data.
The second setting applies GRPO to questions selected by the verifiability classifier, whose compact answers support reliable rule-based rewards.

We use a common GRPO optimization and rollout setup across model families. Runs train for one epoch with BF16, per-device batch size 1, 16 gradient-accumulation steps, learning rate $10^{-6}$ with cosine decay, maximum gradient norm 1, and KL coefficient 0.001. For each prompt, GRPO samples eight completions with temperature 1.0, top-$p$ 0.95, top-$k$ 64, a maximum prompt length of 2,048 tokens, and a maximum completion length of 8,192 tokens. The hard-example branch starts from the corresponding self-distilled SFT checkpoint and uses a GLM-5.2 judge to assign a binary reward by comparing the parsed final answer with the reference; the verifiable-task branch instead uses compact rule-based rewards. Truncated completions are masked from the loss.

\section{Experiments}
\label{sec:experiments}

We organize the results as experiment chains that connect each adapted checkpoint to its starting model and, where applicable, to its immediate SFT parent. This structure makes the effects of ordinary SFT, self-distilled SFT, parameter merging, and GRPO directly interpretable across model families and scales.

\subsection{Evaluation Setup}
\label{sec:evaluation}

We evaluate the target language models on a fixed 778-question subset of FINESSE-Bench~\cite{stanishevskii2026finesse}. Benchmark construction was outside the scope of this work, so we describe only the evaluation set and scoring procedure needed to interpret our model comparisons.

The evaluation set contains 117 CFA Level I questions, 110 CFA Level II questions, 117 CMT Level II questions, 109 VLigaBench-ru problems, 96 technical-analysis trading questions, 115 derivatives-trading questions, and 114 FinQA questions~\cite{chen2021finqa}. We report its micro-accuracy over all 778 questions as the FINESSE-Bench score.

Candidate answers are scored by GPT-5.2 as a binary correctness judge using the question, reference answer, and model response. The retained evaluation configuration uses temperature 0 and a maximum judge response length of 32,768 tokens.

\subsection{Experimental Questions}
\begin{enumerate}[leftmargin=1.7em]
    \item How much does finance-only SFT improve FINESSE-Bench, and which financial components regress?
    \item Does self-distillation preserve the start model's financial behavior better than ordinary SFT at a comparable finance gain?
    \item Can parameter merging recover financial performance lost during adaptation without discarding the domain gain?
    \item Does GRPO improve difficult or verifiable financial reasoning beyond the selected SFT/merged checkpoint?
\end{enumerate}

\subsection{Main Results}
For each selected run, we report the model family, adaptation method, direct start checkpoint, FINESSE-Bench accuracy, and changes relative to both the original model and the direct parent. Within each experiment chain, all rows use the same evaluation set and retained model-specific reasoning configuration.

\par\medskip
\refstepcounter{table}\label{tab:main-results}
\noindent\textbf{Table \thetable:} FINESSE-Bench results for the selected SFT, self-distillation, merging, and RL experiment chains.
\par\smallskip
{\small
\noindent
\begin{tabularx}{\linewidth}{@{}l l X r r r@{}}
\toprule
Model & Adaptation & Start checkpoint & FINESSE & $\Delta$ start & $\Delta$ parent \\
\midrule
\multirow{4}{*}{Qwen3.5-4B}
  & -- & -- & 75.2 & -- & -- \\
  & SFT & Qwen3.5-4B & 72.0 & \textcolor{red!70!black}{-3.2} & \textcolor{red!70!black}{-3.2} \\
  & Self-distilled SFT & Qwen3.5-4B & \textbf{78.0} & \textbf{+2.8} & \textbf{+2.8} \\
  & Linear merge, $w=0.5$ & Qwen3.5-4B SFT & \textbf{76.1} & \textbf{+0.9} & \textbf{+3.0} \\
\midrule
\multirow{2}{*}{Qwen3.5-35B-A3B}
  & -- & -- & 84.2 & -- & -- \\
  & Self-distilled SFT & Qwen3.5-35B-A3B & \textbf{85.2} & \textbf{+1.0} & \textbf{+1.0} \\
\midrule
\multirow{4}{*}{Gemma-4-12B-IT}
  & -- & -- & 77.5 & -- & -- \\
  & SFT & Gemma-4-12B-IT & 73.5 & \textcolor{red!70!black}{-4.0} & \textcolor{red!70!black}{-4.0} \\
  & Self-distilled SFT & Gemma-4-12B-IT & 79.2 & +1.7 & +1.7 \\
  & Hard-sample GRPO & Self-distilled SFT & \textbf{79.6} & \textbf{+2.1} & \textbf{+0.4} \\
\midrule
\multirow{2}{*}{Qwen3-1.7B}
  & -- & -- & 56.0 & -- & -- \\
  & Verifiable-task GRPO & Qwen3-1.7B & \textbf{59.0} & \textbf{+3.0} & \textbf{+3.0} \\
\midrule
\multirow{2}{*}{Qwen3-4B}
  & -- & -- & 61.0 & -- & -- \\
  & Verifiable-task GRPO & Qwen3-4B & \textbf{64.0} & \textbf{+3.0} & \textbf{+3.0} \\
\bottomrule
\end{tabularx}
}
\par\medskip

\subsection{Experiment Design}
To isolate the effect of self-distillation, we compare ordinary and self-distilled SFT using the shared optimization recipe described in Section~\ref{sec:training}. We then compare the adapted checkpoint before and after interpolation with its starting model. Across model families, GRPO uses the same optimization and rollout setup; the hard-example branch uses a binary LLM-judge reward, whereas the verifiable-task branch uses compact rule-based rewards.

\paragraph{GRPO on post-SFT hard examples.}
Starting from the selected self-distilled Gemma-4-12B-IT checkpoint, difficulty-selected GRPO at step 300 raises FINESSE-Bench accuracy from 79.18 to 79.56, an absolute improvement of 0.39 percentage points. This result supports the use of model-relative failure sampling after SFT: even when the starting checkpoint already exceeds the original Gemma model, focusing RL on questions it still fails in at least three of four attempts produces a further aggregate gain. The component-level effects remain mixed and are reported alongside the aggregate result rather than being hidden by it.

\paragraph{RL on verifiable tasks without SFT.}
We separately apply RL directly to unadapted Qwen3 checkpoints using tasks selected for compact answer verification. FINESSE-Bench increases from 56.0 to 59.0 for Qwen3-1.7B and from 61.0 to 64.0 for Qwen3-4B. The consistent three-point improvements show that verifier-compatible RL can provide a useful learning signal even without a preceding domain-SFT stage.

\paragraph{Merging after SFT.}
We also apply an equal-weight linear parameter merge between Qwen3.5-4B and an SFT checkpoint derived from it. The SFT parent scores 73.14 on FINESSE-Bench, while the merged model reaches 76.09. Merging therefore improves upon its SFT parent by 2.96 percentage points and exceeds the original model by 0.90 points. This shows that post-training merging can recover capabilities lost during SFT while preserving a net gain over the starting checkpoint.

\section{Discussion and Insights}
\label{sec:discussion}

The results support three conclusions about stage-specific filtering, financial-capability retention, and the complementary roles of the training sources.

\paragraph{Filtering criteria should follow the training stage.}
Finance relevance is valuable for broad mining, self-containment protects both supervised and synthetic data quality, and answer verifiability determines whether compact RL rewards are defensible. Collapsing these signals into one quality score would hide operationally important differences.

\paragraph{Domain gain is incomplete without retention.}
Ordinary SFT reduces FINESSE-Bench accuracy by 3.2 points for Qwen3.5-4B and 4.0 points for Gemma-4-12B-IT, whereas self-distilled SFT improves the same model families by 2.8 and 1.7 points, respectively. Equal-weight merging recovers 3.0 points over the Qwen SFT parent and finishes 0.9 points above the original model. Both methods mitigate SFT-induced regression, but self-distillation produces the larger gain over the starting checkpoint in the reported Qwen3.5-4B comparison.

\paragraph{Synthetic scale and reasoning diversity are complementary.}
GraphGen supplies controlled concept coverage, mined Glaive examples supply diverse long trajectories, and distilled Finance-Instruct supplies a smaller curated bridge to instruction-following behavior. These sources address different data limitations and form the supervised-training mixture used in the reported experiments.

\section{Conclusion}

We described a data-centric pipeline for financial reasoning post-training built around three explicit curation decisions: finance relevance, self-containment, and compact answer verifiability. These decisions support complementary data paths spanning large-scale reasoning mining, Finance-Instruct distillation, and knowledge-graph-guided synthesis from financial educational material. Across the selected FINESSE-Bench comparisons, ordinary SFT reduces accuracy by 3.2--4.0 points, whereas self-distilled SFT improves over the corresponding starting models by 1.0--2.8 points. Equal-weight merging raises the Qwen3.5-4B SFT checkpoint by 3.0 points and finishes 0.9 points above the original model. GRPO adds 0.4 points after self-distilled SFT on hard examples and 3.0 points when applied directly to verifiable tasks. These results show that explicit data selection, self-distillation, and targeted RL can improve financial reasoning without accepting the financial-capability regressions observed under ordinary SFT.

\nocite{khoroshilov2026quantcode}
\bibliographystyle{plain}
\bibliography{references}

\end{document}